\pdfoutput=1

\documentclass[11pt]{article}

\usepackage{acl}

\usepackage{times}
\usepackage{latexsym}
\usepackage{algorithm}
\usepackage{algorithmic}
\usepackage{graphicx}
\usepackage{amsfonts}
\usepackage{amsmath}
\usepackage{multirow}
\usepackage{multicol}
\usepackage{cases}
\usepackage{makecell}
\usepackage{subcaption}
\usepackage{booktabs}

\newcommand{\thickhline}{\noalign{\hrule height 1pt}}

\usepackage[T1]{fontenc}

\usepackage[utf8]{inputenc}

\usepackage{microtype}

\title{There Is No Standard Answer:  Knowledge-Grounded Dialogue Generation with Adversarial Activated Multi-Reference Learning}

\author{\\
	\textbf{Xueliang Zhao\textsuperscript{1}\footnotemark[2],  Tingchen Fu\textsuperscript{2}\footnotemark[2], Chongyang Tao\textsuperscript{3}, Rui Yan\textsuperscript{2}\footnotemark[1]}\\
	\textsuperscript{1}Wangxuan Institute of Computer Technology, Peking University\\
	\textsuperscript{2}Gaoling School of Artificial Intelligence, Renmin University of China \\
	\textsuperscript{3}Microsoft Corporation\\
	\texttt{\{zhaoxlpku,lucas.futingchen,chongyangtao\}@gmail.com}\\ \texttt{ruiyan@ruc.edu.cn}
}

\begin{document}
\maketitle

\renewcommand{\thefootnote}{\fnsymbol{footnote}}
\footnotetext[2]{The first two authors contribute equally. Xueliang Zhao is responsible for the design of the methodology and algorithm. Tingchen Fu is responsible for the implementation and experiment. The order is decided by a coin flip.}
\footnotetext[1]{Corresponding author: Rui Yan (ruiyan@ruc.edu.cn).}
\setcounter{footnote}{0}
\renewcommand{\thefootnote}{\arabic{footnote}}

\begin{abstract}
Knowledge-grounded conversation~(KGC) shows excellent potential to deliver an engaging and informative response. However, existing approaches emphasize selecting one golden knowledge given a particular dialogue context, overlooking the one-to-many phenomenon in dialogue. As a result, the existing paradigm limits the diversity of knowledge selection and generation. To this end, we establish a multi-reference KGC dataset and propose a series of metrics to systematically assess the one-to-many efficacy of existing KGC models. Furthermore, to extend the hypothesis space of knowledge selection to enhance the mapping relationship between multiple knowledge and multiple responses, we devise a span-based variational model and optimize the model in a wake-sleep style with an ameliorated evidence lower bound objective to learn the one-to-many generalization. Both automatic and human evaluations demonstrate the efficacy of our approach. 

\end{abstract}

\section{Introduction}

Maintaining appropriate human-computer dialogue is an important task leaping toward advanced artificial intelligence and external knowledge is a key ingredient to engaging and meaningful responses~\cite{dinan2018wizard}. To this end, the research area of knowledge-grounded conversation (KGC) has been explored with great interest. In recent years, a number of methods~\cite{lian2019learning,kim2020sequential,zhao2020low,zhao2020knowledge} and benchmarks~\cite{dinan2018wizard,zhou2018dataset} have been proposed. These methods mainly follow the two-step paradigm proposed by~\citet{dinan2018wizard}: Given a dialogue context and a candidate knowledge pool, they (1) first select one or more knowledge passages from the candidate pool, and then (2) generate a response based on the dialogue context and the selected knowledge.

\begin{figure}[t]
 \centering
 \includegraphics[width=0.9\linewidth]{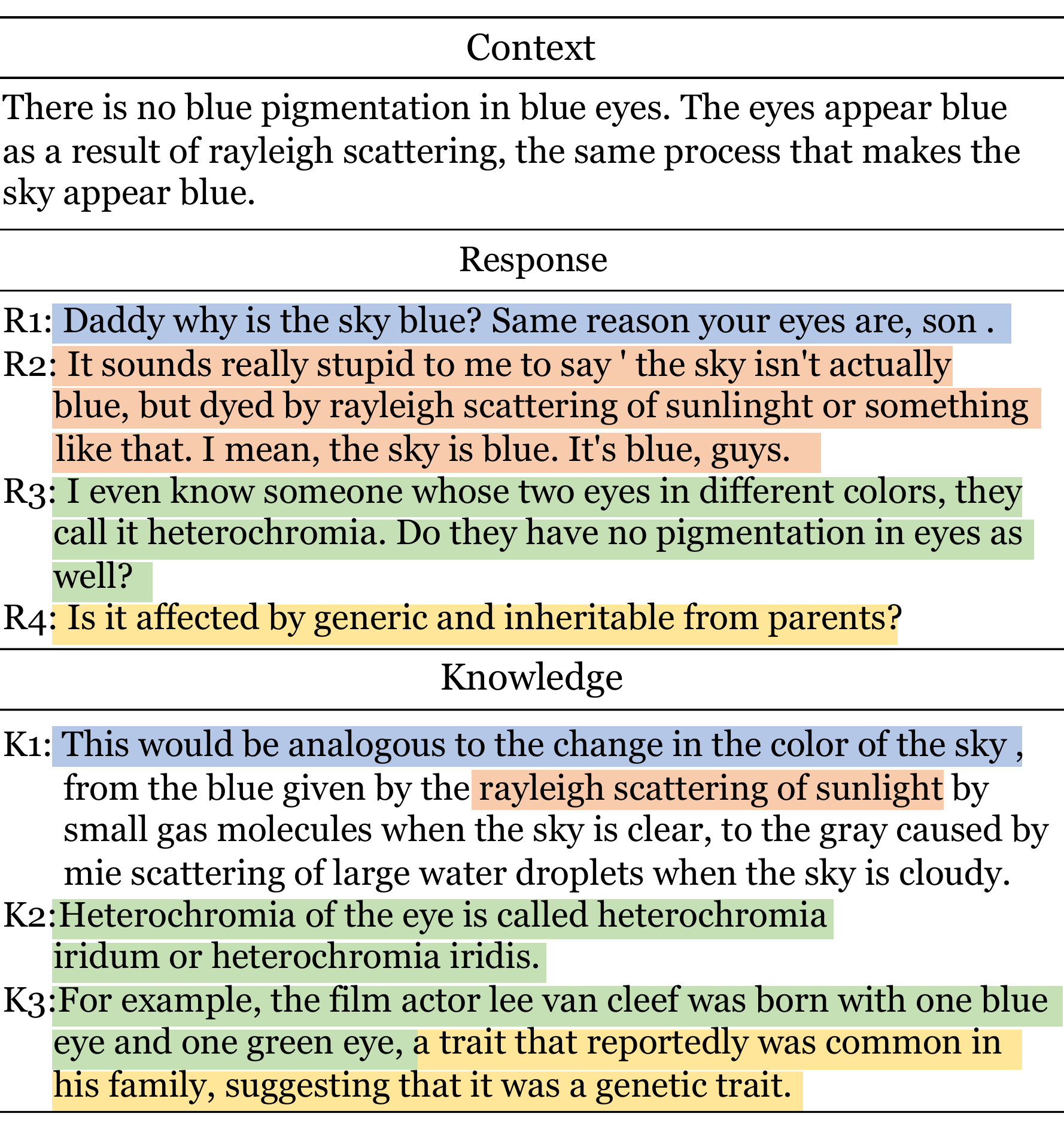} 
 \caption{A conversation from Reddit. Text highlighted in the same color are responses and their corresponding groundings in the knowledge pool.}
 \label{fig:intro_case}
\end{figure}

A large body of works put the emphasis on discovering the \textit{golden knowledge} from the knowledge pool. To be more specific, although many knowledge passages in the candidate pool are relevant to the current conversation context~(context-relevant), usually only one of them pertains to the observed response~(label-relevant), which is often dubbed as golden knowledge by a number of works and researchers. Although many techniques have been developed to discriminate the golden knowledge from the candidate pool, their precision is still far from satisfactory~\cite{zhao2020knowledge}. Moreover, it seems that even humans are unable to accurately identify the so-called golden knowledge.\footnote{According to experiments in \citet{kim2020sequential}, humans could only achieve a precision of 17\% on Wizard of Wikipedia dataset.}
 
In light of the poor performance of humans, we postulate that the so-called golden knowledge is an oversimplification of KGC. Concretely, dialogue is one-to-many in nature with high entropy~\cite{paranjape2022hindsight}, thus there might exist more than one proper knowledge to ground on.
Take a conversation from Reddit as an example (Figure~\ref{fig:intro_case}). All the knowledge is relevant and the four responses grounded on them are reasonable. In a word, there is no such golden knowledge in this case. The hypothesis of golden knowledge overlooks the one-to-many properties in conversation, penalizing perfectly valid knowledge and therefore is harmful to the diversity of generation.

We identify two limitations for previous methods to go beyond the golden knowledge and learn the one-to-many generalization. Firstly, previous methods that tacitly assume the existence of golden knowledge already produce acceptable performance successfully, since most benchmarks~\cite{zhou2018dataset,dinan2018wizard} provide only one response, which coincidentally support the golden knowledge hypothesis when evaluation. Besides, a KGC model has no chance to be exposed to more than one response when training on these benchmarks.
In a word, existing benchmarks are unable to train or evaluate the one-to-many generalization of a model. Second, the golden knowledge is flexible in granularity, not limited to a complete sentence~(Figure~\ref{fig:intro_case}). But previous methods usually limit the granularity of grounding to a complete sentence. Consequently, their decision space of knowledge selection is severely skewed and overfitted by the observed response. In the compressed decision space, they are incapable to model the underlying relationship between the multiple responses and their groundings as well.

In this work, we propose a new KGC framework that is better in one-to-many generalization ability on two counts: 
(1) To train and evaluate the one-to-many generalization ability of a KGC model, we establish the first multi-reference KGC dataset and a series of metrics. 
(2) To extend the hypothesis space of knowledge selection, instead of choosing a knowledge sentence from the candidate set, we design a variational span reading model which directly reads the knowledge text and samples a span as our grounding. We further propose a wake-sleep style learning algorithm to adapt the original evidence lower bound objective~(ELBO) to the multi-reference scenario.
We conduct extensive experiments and both automatic evaluation and human evaluation suggest the efficacy of our methods in multi-reference KGC.

Our contributions are summarized below:

$\bullet$ To our best knowledge, we are the first to explore the one-to-many problem in KGC and establish a multi-reference KGC dataset as well as a series of metrics. 

$\bullet$ We propose a variational span reading model, which reads and comprehends knowledge at a finer granularity and sample a span as the knowledge to ground on.  

$\bullet$ We propose an adversarial activated multi-reference learning algorithm to ameliorate the original ELBO in the multi-reference scenario.

\section{Related Work}
Our work is in line with the research of \textbf{knowledge-grounded conversation}, whose goal is to generate informative responses with external knowledge~\cite{dinan2018wizard,kim2020sequential,zhao2020knowledge}. Since existing benchmarks usually only contain one reference for a conversation~\cite{zhou2018dataset,dinan2018wizard,gopalakrishnan2019topical,wu2019proactive}, most previous works take the assumption of golden knowledge~\cite{zhao2020knowledge,dinan2018wizard}, and some of them use hindsight information from response to detect the golden knowledge~\cite{chen020bridging,kim2020sequential,paranjape2022hindsight}, omitting all the other unobserved but plausible responses. Besides, the granularity of grounding is limited to a complete sentence or passage. Recently, some researchers have attempted to explore the possibility of grounding dialogue with span~\cite{wu2021controllable,meng2020refnet,zhan2021colv}. Their spans are \textit{deterministic} from hard selection process. Differently, we view the span prediction as a \textit{probabilistic} process and propose a variational method to capture the attention span.

\begin{figure*}
    \centering
    \includegraphics[width=0.9\textwidth]{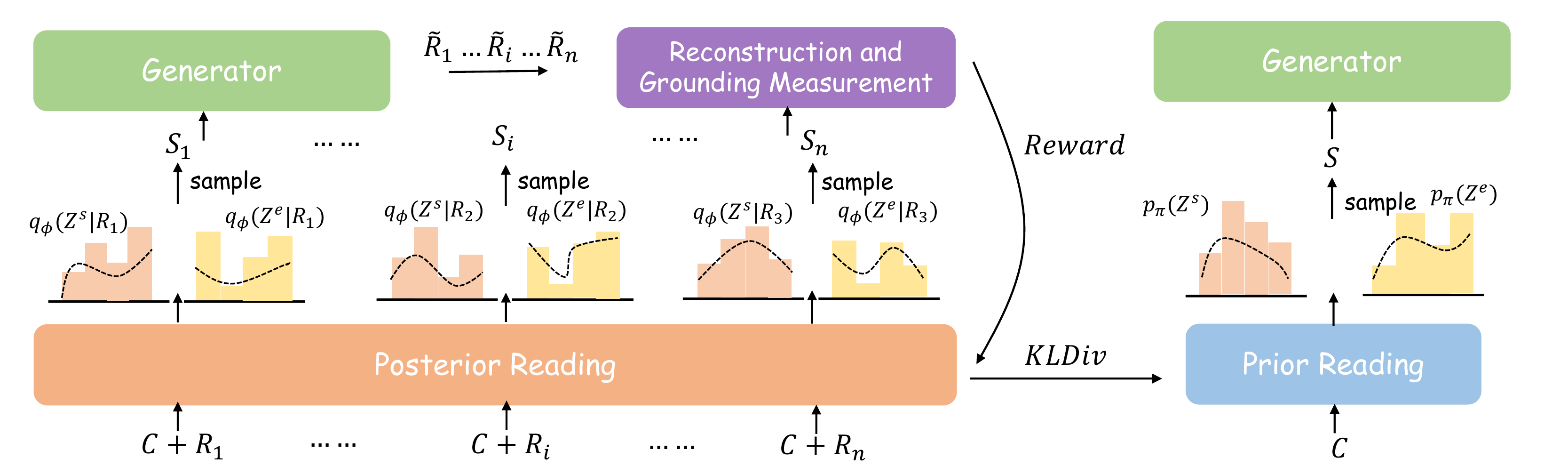}
    \caption{The architecture of the proposed model.}
    \label{fig:architecture}
\end{figure*}

The proposed model also relates to the \textbf{one-to-many} property in dialogue, referring to the phenomenon that the multiple responses are proper for a single dialogue context. How to train and evaluate the one-to-many generalization of a dialogue system is a widely studied topic in open-domain response generation~\cite{gupta2019investigating,zhao2017learning,chan2021enhancing}. Inspired by the efficacy of Variational Auto-Encoder~(VAE), some previous works resort to latent variables to model the one-to-many property of dialogue. For example, \citet{zhao2017learning} model discourse-level diversity with a latent variable subjecting to the Gaussian distribution. \citet{qiu2019training} posit a two-stage method that represents the distinct features of multiple references with a continuous latent variable. However, their latent variables are poor in interpretability. \citet{bao2020plato} and \citet{bao2021plato2} introduce discrete latent variables into the pre-training process. Each value of the latent variable corresponds to the particular latent speech act. As for the evaluation of dialogue system, \citet{gupta2019investigating} show that multi-reference evaluation achieves better correlation with human judgments and release a test set for open-domain dialogue. But to our best knowledge, although \citet{moghe2018towards} construct a multi-reference test set for KGC, there is no standard benchmark for one-to-many training and evaluation in KGC.

\section{Methodology}
\label{sec:preliminary}
\subsection{Problem Formulation and Overview}

For a multi-reference KGC dataset, each case is a triplet $(C, K, \mathcal{R})$ where $C=[w_1,w_2,\cdots,w_{l_{C}}]$ is the context of a conversation composed of previous utterance tokens and $K=[k_1,k_2,\cdots,k_{l_K}]$ is the concatenated sequence of background knowledge and facts. We use $w_i$ and $k_j$ to denote the $i$-th token in context and the $j$-th token in knowledge respectively. $\mathcal{R}=\{R_i\}_{i=1}^n$ is a set of observed responses.
Our goal is to predict various spans $(S_1,S_2,\cdots,S_n)$ in knowledge indicated by the start position $Z^s$ and the end position $Z^e$, and then generate multiple diverse responses $(R_1,R_2,\cdots,R_n)$ accordingly.

The architecture of our approach is exhibited in Figure~\ref{fig:architecture}. It mainly consists of two parts, selective reading (Section~\ref{sec:reading}) and multi-reference learning (Section~\ref{sec:alg}). 
Concretely, for selective reading, we calculate the prior distribution of $Z^s$ and $Z^e$ with the dialogue context and the knowledge, which we refer to as $p_\theta(Z^s)$ and $p_\theta(Z^e)$. The two distributions are used to estimate the joint distribution $p_\theta(Z^s,Z^e)$. Meanwhile, we compute an auxiliary posterior distribution $q_\phi(Z^s|R)$ and $q_\phi(Z^e|R)$, which are used for teaching the prior through minimizing KL-divergence. Note that the posterior is only involved in the training process.
For multi-reference learning, We devise a wake-sleep style learning algorithm. In the wake step, the posterior and generator learn to maximize the evidence lower bound objective with respect to the augmented response set; In the sleep step, a discriminator is trained to distinguish the observed real responses and augmented responses. The two steps are conducted iteratively to learn one-to-many generalization in dialogue.

\subsection{Variational Span Reading}
\label{sec:reading}

\paragraph{Prior Reading.}
To compute the prior distribution of the span, we first concatenate the context and the knowledge together into a single sequence:
\begin{equation}
\mathcal{I}^{pri}=\{w_1,w_2,\cdots,w_{l_C}, k_1,k_2,\cdots,k_{l_K}\},
\end{equation}
before passing through multiple BERT layers~\cite{devlin2019bert}: 
\begin{equation}
\begin{aligned}
 \mathbf{H}^{pri} &= \mathrm{BERT}(\mathcal{I}^{pri}) \in \mathbb{R}^{(l_C+l_K)\times d}.\\
\end{aligned}
\end{equation}
Compared with independent encoding, it allows more sufficient interaction between the dialogue context and knowledge to obtain the context-aware knowledge $\mathbf{K}^{pri}=\mathbf{H}^{pri}_{[l_C:l_C+l_K]}$ as a slice of knowledge part in $\mathbf{H}^{pri}$ and knowledge-aware context representation as a mean pooling of the context part: 
\begin{equation}
    \mathbf{h}^c=\frac{1}{l_C}\sum_{i=1}^{l_C}   \mathbf{H}^{pri}_i.
\end{equation}
 
Next we calculate the joint distribution of $p_\theta(Z^s,Z^e)$. It is not straightforward since it requires enumerating all possibilities of different $Z^s$ and $Z^e$. So we propose to first calculate the distribution of the start position and the end position independently:
\begin{equation}
    \begin{aligned}
    p_\theta(Z^s)={\rm{softmax}}({\rm{MLP}}([\mathbf{h}^c;\mathbf{K}^{pri}])),
    \end{aligned}
\end{equation}
where $\rm{MLP}$ is a multi-layer perceptron. We use $[\cdot;\cdot]$ to denote vector concatenation.\footnote{$\mathbf{K}^{pri}$ is a $l_K \times d$ matrix and $\mathbf{h}^c$ is concatenated to every row of $\mathbf{K}^{pri}$ so $[\mathbf{h}^c;\mathbf{K}^{pri}] \in \mathbb{R}^{l_K \times 2d} $.} $p_\theta(Z^e)$ is calculated in a similar way. Next, we approach the conditional distribution $p_\theta(Z^e|Z^s)$ by aggregating the probability in a constrained area such that the end position always falls behind the start position to form a well-defined span:
\begin{equation}
    \hat{p}_\theta(Z^e = i|Z^s)=
\begin{cases}
\frac{p_\theta(Z^e = i)}{\sum_{j=Z^s}^{l_K}p_\theta(Z^e = j)},& i\geq Z^s \\  
0,& i < Z^s
\end{cases}, 
\end{equation}
thus the join distribution could be efficiently computed as  $p_\theta(Z^s,Z^e)=p_\theta(Z^s)\hat{p}_\theta(Z^e|Z^s)$.

\paragraph{Posterior Reading.}
The hint in a response $R$ is used to identify the latent $Z^s$ and $Z^e$ and calculate $q_\phi(Z^s|R)$ and $q_\phi(Z^e|R)$, which are much easier since the response is a semantic reflection of the span. Firstly, the response is concatenated after the context:
\begin{equation}
\begin{aligned}
\mathcal{I}^{post}=\{&w_1,\cdots,w_{l_C}     r_1,\cdots,r_{l_R}  k_1,\cdots,k_{l_K}\}.
\end{aligned}
\end{equation}
Then the sequence passes through a $3$-layer transformer $\mathcal{F}$:
\begin{equation}
\mathbf{H}^{post}=\mathcal{F}(\mathcal{I}^{post}) \in \mathbb{R}^{(l_C+l_R+l_K) \times d}. 
\end{equation}
Similar to prior reading, the response-aware context representation is pooled with average pooling:
\begin{equation}
\small
\begin{aligned}
\mathbf{h}^{cr}= \frac{1}{l_C+l_R}\sum\limits_{i=1}^{l_C+l_R} \mathbf{H}^{post}_i,
\end{aligned}
\end{equation}
and knowledge representation $\mathbf{K}^{post}$ is the slice of $\mathbf{H}^{post}$ corresponding to the knowledge part.

The hint in the response is sufficient to determine the start point and the end point independently. Thanks to the mean-field approximation, the joint distribution could be factorized as:
\begin{equation}
q_\phi(Z^s,Z^e)=q_\phi(Z^s)q_\phi(Z^e).
\end{equation}
The posterior distribution is calculated as:
\begin{equation}
\begin{aligned}
q_\phi(Z^s|R)={\rm{softmax}}({\rm{MLP}}([\mathbf{h}^{cr};\mathbf{K}^{post}])),
\end{aligned}
\end{equation}
and $q_\phi(Z^e|R)$ is calculated in a similar way.

\paragraph{Generator.}
After obtaining the joint distribution $p_\theta(Z^s,Z^e)$, a sampling from the joint distribution produces a pair of $(Z^s, Z^e)$, corresponding to a span $S$ in knowledge:
\begin{equation}
\label{eq:constitute span}
\begin{aligned}
&(Z^s,Z^e)\sim p_\theta(Z^s,Z^e) \\
&S=[k_{Z^s},k_{Z^s+1},k_{Z^s+2},\cdots,k_{Z^e}].
\end{aligned}
\end{equation}
The sampled span and context are then fed into a generator to predict the response in an auto-regressive way:
\begin{equation}
\small
\begin{aligned}
p_\theta(R|C,Z^s,Z^e)&=p_\theta(R|C,S)=\prod\limits_{t=1} p_\theta(r_t|C,S,r_{<t}).
\end{aligned}
\end{equation}
Theoretically, the generator could be specified as any large-scale pre-trained language model. Here we use GPT-2~\cite{radford2019language}. 
Repeating the sampling process produces multiple diverse spans, thus allowing the generator to synthesize diverse responses for a single case.

\subsection{Adversarial Activated Multi-reference Learning}
\label{sec:alg}
Directly optimizing the marginal likelihood is prohibitively time-consuming and a traditional substitution for marginal likelihood is the evidence lower bound objective~(ELBO):
\begin{equation}
\begin{aligned}
    &\mathcal{L}_{elbo} = \mathbb{E}_{R\in \mathcal{R}} \mathbb{E}_{q_\phi(Z^s,Z^e|R)} \log p(R|C,Z^s,Z^e)\\ &-{\rm{KL}}(q_\phi(Z^s,Z^e|R)||p_\theta(Z^s,Z^e)).
\end{aligned}
\end{equation}
A step-wise derivation could be found in Appendix~\ref{sec:elbo}. After a closer look at the ELBO, we find that the objective is still based on existing responses in $\mathcal{R}$ and tries to maximize the overall likelihood of all the observed data $(C,K,\mathcal{R})$ in our dataset. But as a matter of fact, the one-to-many property of dialogue indicates that the possible responses could be infinite and not enumerable. And a dialogue system is supposed to infer the unobserved responses based on the observed ones, or in other words, be able to discriminate whether a candidate is a possible response or not. In light of this, drawing inspiration from~\citet{hu2018on}, we propose an \underline{A}dversarial \underline{A}ctivated ELBO~(AAELBO):
\begin{equation}
    \begin{aligned}
        &\mathcal{L}_{aaelbo}\\
        &= \mathbb{E}_{R\in \mathcal{R}^A} \mathbb{E}_{q_\phi(Z^s,Z^e|R) d_\pi(R) } \log p(R|C,Z^s,Z^e) \\
        &-{\rm{KL}}(q_\phi(Z^s,Z^e|R)d_\pi(R)||p_\theta(Z^s,Z^e)d(R)),
    \end{aligned}
\end{equation}
where $\mathcal{R}^A$ is the augmented response set comprised of the originally observed response $\mathcal{R}$ and the augmented ones:
\begin{equation}
    \mathcal{R}^A=\mathcal{R} \cup \{R^{aug}_i\}_{i=1}^{\lambda},
\end{equation}
where $\lambda$ is a hyper-parameter. $d_\pi(\cdot)$ is a discriminator with a trainable parameter $\pi$ to classify whether a response is an observed one or an augmented one. $d(\cdot)$ is the corresponding prior defined as the sampling probability among original $\mathcal{R}$ and $\{R^{aug}_i\}_{i=1}^{\lambda}$.
AAELBO is optimized iteratively in two steps:

\paragraph{Sleep-Step.} The parameter of posterior reading, prior reading and generator are fixed. To synthesize $\{R^{aug}_i\}_{i=1}^{\lambda}$, we first calculate the posterior distribution $p_\theta(Z^s,Z^e|R)$ and sample multiple grounding spans accordingly. Then we concatenate the spans to the context respectively and send them to the generator to obtain $\mathcal{R}^A$. The discriminator is a $L$-layer bidirectional transformer trained to maximize the following objective:
\begin{equation}
\small
    \begin{aligned}
        \max\limits_{\pi} \mathbb{E}_{R \in \mathcal{R}^A} \left[y\log d_\pi(R) + (1-y) \log (1-d_\pi(R)) \right],
    \end{aligned}
\end{equation}
where $y=1$ if $R$ is an observed response in the original dataset else $y=0$.

\paragraph{Wake-Step.} The parameter of the discriminator is frozen. We use the discriminator to assign an importance weight to each response in the candidate set. The posterior reading, prior reading and the generator are then optimized on the augmented $\mathcal{R}^A$ with the importance weight given by the discriminator. Mathematically, the training objective in this step is:
\begin{equation}
    \begin{aligned}
        &\max\limits_{\theta,\phi} \mathbb{E}_{R\in \mathcal{R}^A} \mathbb{E}_{q_\phi(Z^s,Z^e|R) d_\pi(R) } \log p(R|C,Z^s,Z^e) \\
        &-{\rm{KL}}(q_\phi(Z^s,Z^e|R)d_\pi(R)||p_\theta(Z^s,Z^e)d(R)).
    \end{aligned}
\end{equation}
As spans are obtained from a discrete sampling process, the gradient of the AAELBO objective is not differentiable to $\phi$. Therefore, we exploit policy-gradient method~\cite{sutton2000policy} to estimate the gradient, which is formulated as:
\begin{equation}
\begin{aligned}
\label{eq:update phi}
    &\nabla_\phi \mathbb{E}_{R\in \mathcal{R}^A}  \mathbb{E}_{q_\phi(Z^{s}, Z^{e}) d_\pi(R)} \log p(R|C,Z^s,Z^e)\\
    =&\mathbb{E}_{R\in \mathcal{R}^A} \mathbb{E}_{q_\phi(Z^{s}, Z^{e})d_\pi(R)} \nabla_\phi \log q_\phi(Z^s,Z^e) {\rm{Re}}.
\end{aligned}
\end{equation}
Conventionally, the reward ${\rm{Re}}$ is calculated as $\log p(R|C,Z^s,Z^e)$ in a teacher-forcing way, which is incompetent in modeling the complex mapping relationship between the multiple spans and multiple references. 
As a possible remedy, we propose to reinforce the relationship between the pairs of span and the response for both the posterior reading and the generation. We ameliorate the original reward to adapt to the multi-reference scenario:
\begin{equation}
\label{eq:new reward}
\small
{\rm{Re}}=d_\pi(R)\left(\alpha{\rm{Rec}}\left(R^{gen},\mathcal{R}^A\right) + {\rm{Gnd}}\left(S,R^{gen}\right)\right).
\end{equation}
The reward is composed of two parts: the reconstruction reward  ${\rm{Rec}}(R^{gen},\mathcal{R}^A)$ and the grounding reward ${\rm{Gnd}}(S,R^{gen})$, which we will elaborate as below. To optimize the objective, we first sample a response $R \in \mathcal{R}^A$ and calculate $q_{\phi}(Z^s,Z^e|R)$. Next, we sample a span $S$ and send the span to the generator together with the context to synthesize $R^{gen}$. $\alpha$ is a hyper-parameter.

\paragraph{Reconstruction Reward.}
The reconstruction reward is designed for strengthening the span-response mapping relationship in posterior reading:
\begin{equation}
\begin{aligned}
    {\rm{Rec}}(R_{gen},\mathcal{R}^A)&= \frac{1}{|\mathcal{R}^A|} \sum\limits_{i=1}^{|\mathcal{R}^A|} 
    \left(y_i s(R_{gen},R_i) \right.\\ 
    &\left. +\left(1-y_i\right)\left(1-s(R_{gen}, R_i)\right)\right).
\end{aligned}
\end{equation}
We have $y_i=1$ if $R_i$ is the sampled response $R$ else $0$. $s(\cdot,\cdot)$ is a similarity function. The reconstruction reward gives the posterior a bigger reward when the span predicted by the posterior is easy for the generator to synthesize the corresponding response.

\paragraph{Grounding Reward.}
The grounding reward is designed for strengthening the span-response mapping relationship in generation. 
It uses BERT as its backbone, accepts a span and a generated response as input:
\begin{equation}
    \mathcal{I}_{gnd}=\{r_1,\cdots,r_{l_R}  k_{Z^s},\cdots,k_{Z^e}\},
\end{equation}
and maps the representation of the [CLS] token to a grounding reward:
\begin{equation}
\begin{aligned}
    \mathbf{H}&=\operatorname{BERT}(\mathcal{I}_{gnd}), \\
    {\rm{Gnd}}(S_i,R_j)&=\sigma({\rm{MLP}}(\mathbf{H}_{[{\rm CLS}]})),
\end{aligned}
\end{equation}
where $\sigma(\cdot)$ denotes Sigmoid function.
To train the discrimination network, for every response $R_i$, we first heuristically tag a corresponding span in knowledge text as a pseudo span label:
\begin{equation}
\bar{S}_i=\mathop{\mathrm{argmax}}\limits_{S \in \Omega} s(S,R),
\end{equation}
where $s(\cdot,\cdot)$ is a similarity function and $\Omega$ is a candidate set constructed by enumerating all the possible spans in the knowledge with a sliding window. 
The grounding reward network is trained to minimize the following objective:
\begin{equation}
\small
\begin{aligned}
    &\mathcal{L}_{gnd}=\frac{1}{|\mathcal{R}|}\sum\limits_{i=1}^{|\mathcal{R}|} \max\{0, \mu + {\rm{Gnd}}(\bar{S_j},R_{i})  - {\rm{Gnd}}(\bar{S_i},R_i)\}\\
    &j\sim {\rm{Uniform}}\{1,2,\cdots,i-1,i+1,\cdots,|\mathcal{R}|\},
\end{aligned}
\end{equation}
where $\mu$ is a hyper-parameter.

\section{Experiment Setup}
\subsection{Dataset}

We establish a multi-reference KGC dataset with conversations from Reddit. As a social news aggregation, conversations in Reddit are well-grounded by an external website, usually a Wikipedia page. Elaborated filtering and cleaning are carried out to construct a multi-reference KGC dataset with a training set, a validation set and two test sets, namely General Test and Focused Test. In the Focused Test, multiple references are grounded within a single knowledge sentence. So it is designed to evaluate the one-to-many generalization ability only with respect to the grounding granularity. Apart from that, we also develop a General Test, in which the grounding on the knowledge passages is unconstrained since in real-world scenarios, it is more common that multiple references are grounded on various knowledge. The statistics of the dataset are shown in Table~\ref{tab:dataset}.
For more details about the data collection protocol, please refer to Appendix~\ref{sec:dataset}.\footnote{Our code and dataset is available at \scriptsize \url{https://github.com/TingchenFu/MultiRefKGC}}

\begin{table}[h!]
\resizebox{1.0\linewidth}{!}{
\begin{tabular}{lcccc}
\toprule
                & Train     & Valid  & \makecell[c]{General \\ Test} & \makecell[c]{Focused \\ Test} \\
\midrule
\# Dialogues    & 113,026   & 7,225  & 9,837        & 833          \\
\# Utterences   & 1,142,616 & 61,088 & 84,682       & 7,578        \\
\# Knowledge    & 1,530,013 & 87,070 & 116,379      & 7,747        \\
\midrule
\textit{AvG.Len}(\# words): & \multicolumn{4}{c}{}                             \\
Utterances      & 33.46     & 32.45  & 32.94        & 32.51        \\
Knowledge       & 30.01     & 30.03  & 29.86        & 30.59    \\
\bottomrule
\end{tabular}
}
\caption{Statistics of the Multi-Reference KGC Dataset.}
\label{tab:dataset}
\end{table}

\subsection{Baselines}
We compare our proposed approach with the following methods: (1) \textbf{MTASK-RF}~\cite{ghazvininejad2018knowledge} is an early model for KGC using an independent dialogue encoder and fact encoder to encode utterance history and knowledge separately. (2) \textbf{TMN}~\cite{dinan2018wizard} is a transformer version of memory network. (3) \textbf{VHRED}$_{lgm}$~\cite{zhao2020multi} is a variational hierarchical model with linear Gaussian prior and is trained with multiple augmented responses.
(4) \textbf{SKT}~\cite{kim2020sequential} uses sequential latent variables to predict the grounding knowledge sentences at each turn. 
(5) \textbf{CGRG}~\cite{wu2021controllable} is a two-stage method that first predicts the control phrase and then generates a response with a GPT-2 that is extended with inductive attention.  (6) \textbf{KnowledGPT}~\cite{zhao2020knowledge} jointly trains a knowledge selector and a generator with the policy-gradient method and curriculum learning, achieving state-of-the-art performance on two benchmarks.
(7) \textbf{KTWM}~\cite{zheng2021knowledge} incorporates term-level denoising into the knowledge selection and generates a simulated response vector to determine the fine-grained weight of knowledge terms. 
(8) \textbf{CoLV}~\cite{zhan2021colv} uses two latent variables to boost the diversity in knowledge selection and response generation, respectively.  
(9) \textbf{K2R}~\cite{adolphs2021reason} is a new method that first probes knowledge from a large-scale pre-trained language model and then generates a response with the context and probed knowledge. 

All baselines are implemented strictly following the official code and the original paper. Their parameters are tuned to achieve the best results on the validation set.

\begin{table*}[!t]
\centering 
\resizebox{0.9\linewidth}{!}{
\begin{tabular}{lccccccccccc}
\toprule
\multicolumn{1}{c}{\multirow{2}{*}{Models}} & \multicolumn{2}{c}{BLEU}                                         & \multicolumn{3}{c}{ROUGE}                       & \multicolumn{2}{c}{Entropy}                                        & \multicolumn{2}{c}{Intra-Dist} & \multicolumn{2}{c}{Inter-Dist} \\
\cmidrule(lr){2-3} \cmidrule(lr){4-6} \cmidrule(lr){7-8} \cmidrule(lr){9-10} \cmidrule(lr){11-12}
\multicolumn{1}{c}{}                        & B-1             & B-2                          & R-1             & R-2             & R-L             & E-1             & E-2                          & D-1             & D-2             & D-1             & D-2             \\ \midrule
MTASK-RF~\cite{ghazvininejad2018knowledge}                                    & 0.228          & 0.081                   & 0.143          & 0.017          & 0.127          & 5.391          & 7.481                   & -              & -              & 0.057          & 0.231          \\ 
TMN~\cite{dinan2018wizard}                                          & 0.234          & 0.066                    & 0.160          & 0.029          & 0.138          & 3.314          & 3.727                     & -              & -              & 0.001          & 0.002          \\ 
SKT~\cite{kim2020sequential}                                          & 0.172          & 0.054                    & 0.140          & 0.026          & 0.123          & 3.433          & 4.335                     & 0.169          & 0.182          & 0.018          & 0.061          \\ 
KnowledGPT~\cite{zhao2020knowledge}                                   & 0.306          & 0.110                    & 0.173          & 0.037          & 0.153          & 5.144          & 7.547                     & -              & -              & 0.047          & 0.201          \\ 
CGRG~\cite{wu2021controllable}                                         & 0.275          & 0.075                    & 0.137          & 0.018          & 0.120          & 5.694          & 8.759           & -              & -              & 0.040          & 0.216          \\ 
VHRED$_{lgm}$~\cite{zhao2020multi}                                        & 0.226          & 0.086                           & 0.113          & 0.014          & 0.104          & 3.217          & 4.190           & 0.230          & 0.263          & 0.003          & 0.007          \\ 
KTWM~\cite{zheng2021knowledge} & 0.309 &0.108 & 0.142 & 0.020 & 0.126 & 3.118 & 3.688 & -&-& 0.002 & 0.004 \\ 
CoLV~\cite{zhan2021colv}  & 0.308 & 0.116  &0.169  &0.034 & 0.126 & 5.149 & 8.679 & 0.352 & 0.383 & 0.078 & 0.284 \\ 
K2R~\cite{adolphs2021reason}   & 0.299 & 0.109 & 0.174  & 0.033 & 0.152 & 3.168 & 4.936 & - & - & 0.067 & 0.204 \\ \midrule
Ours         & \textbf{0.322}$^{\star}$  & \textbf{0.121}$^{\star}$   & \textbf{0.178}$^{\star}$  & \textbf{0.041} & \textbf{0.156} & \textbf{6.195}$^{\star}$  & \textbf{8.806}$^{\star}$           & \textbf{0.397}$^{\star}$  & \textbf{0.487}$^{\star}$  & \textbf{0.091}$^{\star}$  & \textbf{0.310}$^{\star}$  \\ 
\bottomrule
\end{tabular}
}
\caption{Automatic evaluation results on the General Test. Numbers in bold are the best results. Significant improvements over the best baseline results are marked with ${\star}$
(t-test, $p<0.05$). }
\label{tab:exp_gen}
\end{table*}

\begin{table*}[!t]
\centering
\resizebox{0.9\linewidth}{!}{
\begin{tabular}{lccccccccccc}
\toprule
\multicolumn{1}{c}{\multirow{2}{*}{Models}} & \multicolumn{2}{c}{BLEU}     & \multicolumn{3}{c}{ROUGE} & \multicolumn{2}{c}{Entropy}  & \multicolumn{2}{c}{Intra-Dist} & \multicolumn{2}{c}{Inter-Dist} \\ 
\cmidrule(lr){2-3} \cmidrule(lr){4-6} \cmidrule(lr){7-8} \cmidrule(lr){9-10} \cmidrule(lr){11-12} 
\multicolumn{1}{c}{}                        & B-1    & B-2       & R-1      & R-2      & R-L     & E-1    & E-2      & D-1             & D-2             & D-1             & D-2             \\ \midrule
MTASK-RF~\cite{ghazvininejad2018knowledge}                                     & 0.116 & 0.034  & 0.122   & 0.013   & 0.111  & 4.756 & 6.274  & -              & -              & 0.108          & 0.311          \\ 
TMN~\cite{dinan2018wizard}                                          & 0.157 & 0.041  & 0.108   & 0.014   & 0.099  & 2.906 & 2.607  & -              & -              & 0.004          & 0.007          \\ 
SKT~\cite{kim2020sequential}                                          & 0.125 & 0.032  & 0.100   & 0.011   & 0.092  & 3.118 & 4.007 & 0.172          & 0.185          & 0.053          & 0.135          \\ 
KnowledGPT~\cite{zhao2020knowledge}                                  & 0.164 & 0.033 & 0.142   & 0.022   & 0.125  & 4.708 & 6.431  & -              & -              & 0.107          & 0.301          \\ 
CGRG~\cite{wu2021controllable}                                         & 0.186 & 0.040 & 0.137   & 0.012   & 0.103  & 5.484 & \textbf{7.949}  & -              & -              & 0.135          & 0.471          \\ 
VHRED$_{lgm}$~\cite{zhao2020multi}                                        & 0.127 & 0.036  & 0.088   & 0.007   & 0.083  & 3.101 & 3.951 & 0.233          & 0.265          & 0.014          & 0.037          \\ 
KTWM~\cite{zheng2021knowledge}   &0.188 & 0.042 & 0.123 & 0.014 & 0.110& 3.126 & 3.689 & - & - & 0.009 & 0.034 \\ 
CoLV~\cite{zhan2021colv} & 0.186 & 0.039 & 0.098 & 0.010 & 0.089 & 5.758& 7.674 & 0.274 & 0.357 & 0.178 & 0.457 \\
K2R~\cite{adolphs2021reason} & 0.184 & 0.041 & 0.135 & 0.023 & 0.121 & 5.766 & 7.826 & - & - & 0.167 & 0.226 \\ \midrule
Ours                                         & \textbf{0.203}$^{\star}$ & \textbf{0.043}  & \textbf{0.149}$^{\star}$   & \textbf{0.028}$^{\star}$   & \textbf{0.129}  & \textbf{5.891}$^{\star}$ & 7.893  & \textbf{0.323}$^{\star}$          & \textbf{0.407}$^{\star}$          & \textbf{0.191}$^{\star}$          & \textbf{0.473}          \\
\bottomrule
\end{tabular}
}
\caption{Automatic evaluation results on the Focused Test. Numbers in bold are the best results. Significant improvements over the best baseline results are marked with ${\star}$
(t-test, $p<0.05$).}
\label{tab:exp_focus}
\end{table*}

\section{End-to-End Evaluation}
\subsection{Evaluation Metrics}
We choose distinct~\cite{li2015diversity}, entropy, BLEU~\cite{papineni2002bleu}\footnote{\scriptsize\url{ftp://jaguar.ncsl.nist.gov/mt/resources/mteval-v14c.pl}} and ROUGE~\cite{lin2004rouge}\footnote{\scriptsize\url{https://github.com/bckim92/language-evaluation}} to be our automatic metrics. 
BLEU and ROUGE evaluate the appropriateness of the proposed model while distinct and entropy focus on the diversity of generation.\footnote{In the rest of this paper, we use B-$i$ as a shorthand for BLEU-$i$, R-$i$ as a shorthand for Rouge-$i$, E-$i$ for Entropy-$i$ and D-$i$ for Distinctness-$i$, $i \in \{1,2\}$}
We measure both Inter-Dist (the distinct score of generated text in the whole dataset) and Intra-Dist (the averaged distinct score of multiple generated hypotheses for every single case)\footnote{Intra-dist is only measured for VAE-based methods as others could not generate multiple responses for one case} following \citet{qiu2019training}.  
Apart from automatic evaluation, $300$ examples are randomly sampled from the General Test and well-educated native speakers are recruited to assess the quality of the generation from different models. Each annotators are required to given a score in $\{0: \mathrm{bad},1:  \mathrm{fair}, 2:\mathrm{good}\}$  for three aspects: (1) \textit{fluency}: whether the reply is fluent; (2) \textit{coherence}: whether the reply is coherent with the context; and (3) \textit{faithfulness}: whether the reply is well-grounded and faithful to the clue. The agreement of annotators is measured via Fleiss' kappa~\cite{fleiss1971measuring}.

\subsection{Results and Discussions}

\begin{table*}[t!]
\centering
\resizebox{0.85\linewidth}{!}{
\begin{tabular}{lcccccccc}
\toprule
\multicolumn{1}{c}{\multirow{2}{*}{Models}} & \multicolumn{4}{c}{General Test}               & \multicolumn{4}{c}{Focused Test}                  \\ \cmidrule(lr){2-5} \cmidrule(lr){6-9} 
\multicolumn{1}{c}{}                        & Fluency & Coherence & Faithfulness & kappa & Fluency & Coherence & Faithfulness & kappa \\ 
\midrule
KnowledGPT                                   & 1.64    & 1.86      & 1.51         & 0.68  & 1.58    & 1.74      & 1.56         & 0.71  \\
CGRG                                         & 1.51    & 1.77      & 1.63         & 0.71   & 1.54    & 1.81      & 1.68         & 0.67  \\
CoLV  & 1.63 & 1.83 & 1.59 & 0.67 & 1.57 & 1.79 & 1.62 & 0.68 \\
K2R  & 1.65  & 1.87 & 1.33 & 0.69 & 1.61 & 1.77 & 1.29 & 0.71 \\ \midrule
Ours                                         & \textbf{1.70}    & \textbf{1.91}      & \textbf{1.74}         & 0.70  & \textbf{1.63}    & \textbf{1.87}      & \textbf{1.77}         & 0.68  \\
\bottomrule
\end{tabular}
}
\vspace{-1mm}
\caption{Human evaluation results. Numbers in bold are the best results.}
\vspace{-1mm}
\label{tab:exp_human}
\end{table*}

\begin{table*}[t!]
\centering \small
\begin{tabular}{lccccccccccc}
\toprule
\multicolumn{1}{c}{\multirow{2}{*}{Model}} & \multicolumn{2}{c}{BLEU} & \multicolumn{3}{c}{ROUGE} & \multicolumn{2}{c}{Entropy} & \multicolumn{2}{c}{Intra-Dist} & \multicolumn{2}{c}{Inter-Dist} \\
\cmidrule(lr){2-3} \cmidrule(lr){4-6} \cmidrule(lr){7-8} \cmidrule(lr){9-10} \cmidrule(lr){11-12} 
                       & B-1          & B-2         & R-1      & R-2     & R-L     & E-1           & E-2           & D-1             & D-2            & D-1             & D-2            \\
\midrule
Ours             & \textbf{0.322$^\star$}       & \textbf{0.121$^\star$}      & \textbf{0.178$^\star$}   & \textbf{0.041}  & \textbf{0.156}  & 6.195        & 8.806        & \textbf{0.397$^\star$}          & \textbf{0.487$^\star$}         & 0.091          & 0.310        \\ 
\textit{-span}                   & 0.302       & 0.115      & 0.171   & 0.039  & 0.152  & 6.238        & 8.889        & 0.351          & 0.417         & 0.090          & 0.314         \\
\textit{-dis}          & 0.321       & 0.118      & 0.170   & 0.039  & 0.147  & 6.057        & 8.510        & 0.196          & 0.218         & 0.075          & 0.257         \\
\textit{-rec}                  & 0.299       & 0.112      & 0.167   & 0.038  & 0.149  & 6.052        & 8.499        & 0.337          & 0.389         & 0.092          & 0.300         \\
\textit{-ground}              & 0.319       & 0.118      & 0.140   & 0.026  & 0.123  & 6.377        & 9.049        & 0.206          & 0.230         & 0.088          & 0.310         \\
\bottomrule
\end{tabular}
\vspace{-2mm}
\caption{Ablation study on the General Test. Numbers in bold are the best results. Significant differences over other variants are marked with $^\star$ (t-test, $p < 0.05$).}
\vspace{-2mm}
\label{tab:ablation}
\end{table*}

The automatic evaluation results on General Test are presented in Table~\ref{tab:exp_gen}. 
We can have the following observations:
(1) Our model outperforms most of the baseline methods in both appropriateness and diversity, especially  KnowledGPT, a competitive baseline in KGC, due to the more flexible grounding. To verify this point, we measure the unigram F1 between the chosen knowledge/span and the corresponding responses. The result is $13.6\%$ for KnowledGPT and $14.4\%$ for ours.  
(2) CGRG and CoLV both achieve comparable distinct scores to ours, thanks to their control phrase and the latent variable in generation, respectively.
(3) KTWM achieves a competitive appropriateness performance due to its fine-grained weight mechanism. But without consideration for the multi-reference scenario, it disproportionately attends to generic words, indicated by its poor diversity. 
(4) SKT and VHRED$_{lgm}$ are VAE-based methods as well. SKT highly relies on ground-truth knowledge labels, which are not always available in KGC datasets. VHRED$_{lgm}$ supports multi-reference training but does not take external knowledge into account. Its poor performance reveals the necessity of a multi-reference KGC model.

The automatic results on the Focused Test are shown in Table~\ref{tab:exp_focus}. 
When comparing Table~\ref{tab:exp_focus} with Table~\ref{tab:exp_gen}, we could see a decline in appropriateness for nearly all methods and a drop or fluctuation in Intra-dist for three VAE-based models. We conjecture the reason is that the case in the Focused Test is much more challenging and their responses are more semantically concentrated. 
The advantage of the proposed model over KnowledGPT is more obvious since KnowledGPT only considers the selection of knowledge at a coarse granularity.
The performance of CGRG is impressive in distinct and entropy with the help of control phrases. Conversely, K2R is low in diversity in both General Test and Focused Test. We gauge that is because of the knowledge module of K2R. It is in vanilla encoder-decoder architecture and is unable to generate diverse knowledge, thus limiting the hypothesis space of generation.

The human evaluation results are displayed in Table~\ref{tab:exp_human}.\footnote{We only conduct a human evaluation on three representative models as human labor is expensive.} 
Notably, there is a significant improvement in \textit{faithfulness} for the proposed model. We attribute this to the span mechanism as the semantics in a span are more concentrated. The kappa signifies the effectiveness of human evaluation. Also, we note that K2R is poor in faithfulness since its knowledge is generated and might suffer from hallucinations.

A case study could be found in Appendix~\ref{sec:case}.

\subsection{Ablation Study}
To understand the impact of each component, we compare the proposed model with several variants on General Test: 
(1) \textit{-span}: the degenerated knowledge sentence selection version of our model.\footnote{In practice, the distribution of the end position $Z^e$ is disabled and we select the knowledge sentence that the sampled $Z^s$ falls in.} 
(2) \textit{-dis}: the discrimination network is removed and our training objective is reduced to vanilla ELBO; 
(3) \textit{-rec}: the Reconstruction reward is removed.
(4) \textit{-ground}: the Grounding reward is removed.
From the results presented in Table~\ref{tab:ablation}, we could observe that 
(1) The removal of the span mechanism causes a drop in appropriateness since the irrelevant parts in a complete knowledge passage bring noise to the model.
(2) The discrimination network plays an important role in improving the diversity of the generation, indicated by the performance of \textit{-dis}. It is reasonable since our AAELBO augments the original response set with unobserved responses.
(3) Both reward components are crucial as the removal of any destroys the mapping relationship between grounding span and response, leading to a sub-optimal solution.

\section{One-to-Many Evaluation}

\subsection{Evaluation Metrics}
One-to-many generalization pays attention to the diversity of a dialogue model. In KGC, the diversity originates from not only the synthesis process of the generator but also the knowledge selection process. Intra-Dist and Inter-Dist only evaluate the diversity of the end-to-end generation, thus are insufficient to measure the effect of each component. Inspired by \citet{shen2019select}, we propose a series of metrics to fill in the gap:
(1) \textbf{The ratio of unique grounding}: When conducting repetitive experiments, the ratio of unique knowledge~(either in the forms of sentence or span) is selected.\footnote{For example, if the knowledge selector selects two different knowledge in 5 repetitive experiments, the ratio of unique grounding is 40\%} This metric measures the diversity in the grounding process. 
(2) \textbf{The ratio of unique generation}: When conducting repetitive experiments, the ratio of unique generated responses. This metric measures overall diversity. 
(3) \textbf{The effect of grounding}: When conducting repetitive experiments, the ratio of unique grounding to unique generation, or the ratio of (2) to (1). It measures the diversity contributed by the generator and the influence of the knowledge.

\begin{table}[t!]
\centering
\resizebox{0.8\linewidth}{!}{
\begin{tabular}{lccc}
\toprule
& \makecell[c]{\%unique \\Grounding}  &\makecell[c]{\%unique \\Generation} & \makecell[c]{Effect of \\Grounding} \\ 
\midrule
SKT  &  0.401 & 0.336 & 0.838  \\
CoLV & 0.679  & 0.332 & 0.488   \\
Ours & \textbf{0.788$^\star$} & \textbf{0.628$^\star$}     & 0.797 \\ 
\midrule
\textit{-span} & 0.392 & 0.359     & \textbf{0.916}$^\star$          \\
\textit{-dis} & 0.596 & 0.318     & 0.533  \\
\bottomrule
\end{tabular}
}
\caption{Diversity analysis on the General Test. Numbers in bold are the best results. Best result with significant difference is marked with ${\star}$ (t-test, $p<0.05$).}
\label{tab:exp_diverse}
\end{table}

\subsection{Results And Discussions}
We choose two VAE-based methods SKT~\cite{kim2020sequential} and CoLV~\cite{zhan2021colv} as our baseline and also include two variants of our approach, namely \textit{-span} and \textit{-dis}. We note that \textit{-span} accomplishes the best result in the effect of span and the result of SKT is very similar. That is because their grounding is always a complete knowledge sentence, thus more influential and decisive when fed into the generator. This also accounts for the low ratio of the unique span since their decision space of knowledge selection is limited. Besides, when comparing \textit{-dis} and CoLV, which is also a span-based method, we could conclude that the latent variables of CoLV help to boost the generation diversity. Our method achieves the best results on the ratio of unique grounding and the effect of grounding, verifying the effectiveness of our proposed AAELBO.

\section{Conclusions}
We have shown that the proposed variational knowledge attention method is helpful to ground a dialogue flexibly at different levels of granularity. Besides, we devise a wake-sleep style learning algorithm to adapt the original ELBO. And to enhance the mapping relationship between different spans and different responses, we ameliorate the original reward in REINFORCE~\cite{williams1992simple} to adapt to the multi-reference scenario. We have demonstrated the efficacy of our model with extensive experiments.

\section*{Limitations}
This paper has presented an approach to address the one-to-many generalization problem in KGC. All technologies built upon the large-scale PLM more or less inherit their potential harms~\cite{bender2021dangers}. Besides, we acknowledge some specific limitations: 

(1) In the dataset collection, we use unigram-F1 to measure the similarity between the response and the knowledge passage. This method is not exactly precise and could miss useful information or introduce unwanted noise. If the selected knowledge is not accurate, the response may contain extra hallucinations. To make up for that, we recruit crowd workers to control the quality of our dataset.

(2) In the generation process, we sample a single span to ground. However, sometimes choosing multiple pieces of knowledge has the potential to include more useful information. If this is required, we could simply sample multiple times (Eq.11) to obtain multiple spans for grounding.

\section*{Ethics Statement}
This paper studies knowledge-grounded conversation. We extend the existing paradigm to the multi-reference scenario, which is more practical in real-world settings. The dataset we constructed contains no personal identifiable information and the proposed approach does not introduce ethical or societal prejudice.

\section*{Acknowledgements}
This work was supported by National Natural Science Foundation of China (NSFC Grant No. 62122089 and No. 61876196), Beijing Outstanding Young Scientist Program NO. BJJWZYJH012019100020098, and Intelligent Social Governance Platform, Major Innovation \& Planning Interdisciplinary Platform for the ``Double-First Class'' Initiative, Renmin University of China. This work was also supported in part by Independent Research Fund Denmark under agreement 8048-00038B. We wish to acknowledge the support provided and contribution made by Public Policy and Decision-making Research Lab of RUC. Rui Yan is supported by Beijing Academy of Artificial Intelligence (BAAI).

\bibliography{anthology,custom}

\appendix
\clearpage

\section{Derivation of ELBO}
\label{sec:elbo}
\begin{equation}
\small
\begin{aligned}
&\log p(\mathcal{R}|C,K)\\
=& \log \sum\limits_{i=1}^{|\mathcal{R}|} p(R_i|\mathcal{R}) \sum\limits_{(Z^s,Z^e)}p(R,Z^s,Z^e|C,K)\\
=&\log \sum\limits_{i=1}^{|\mathcal{R}|}  p(R_i|\mathcal{R}) \sum\limits_{(Z^s,Z^e)}p(R,Z^s,Z^e|C,K)\frac{q_\phi(Z^s,Z^e)}{q_\phi(Z^s,Z^e)}\\
=&\log \sum\limits_{i=1}^{|\mathcal{R}|}  p(R_i|\mathcal{R}) \mathbb{E}_{q_\phi(Z^s,Z^e)} \frac{p(R,Z^s,Z^e|C,K)}{q_\phi(Z^s,Z^e)} \\
\geq& \mathbb{E}_{R \in \mathcal{R}}  \mathbb{E}_{q_\phi(Z^s,Z^e)} \log\frac{p(R,Z^s,Z^e|C,K)}{q_\phi(Z^s,Z^e)}\\
=&\mathbb{E}_{R \in \mathcal{R}}  \mathbb{E}_{q_\phi(Z^s,Z^e)}\log p(R,Z^s,Z^e|C,K)\\
-&\mathbb{E}_{R \in \mathcal{R}} \mathbb{E}_{ q_\phi(Z^s,Z^e)}\log q_\phi(Z^s,Z^e)\\
=&\mathbb{E}_{R \in \mathcal{R}} \mathbb{E}_{ q_\phi(Z^s,Z^e)}\left[\log p(R,Z^s,Z^e|C,K)-\log q_\phi(Z^s,Z^e)\right]\\
=&\mathbb{E}_{R \in \mathcal{R}} \mathbb{E}_{ q_\phi(Z^s,Z^e)}[\log p(R|Z^s,Z^e)+\log p_\pi(Z^e|Z^s)\\
+&\log p_\pi(Z^s)-\log q_\phi(Z^s,Z^e)]
\end{aligned}
\end{equation}
Note that we use $q_\phi(Z^s,Z^e)$ as a shorthand for $q_\phi(Z^s,Z^e|R)$ to avoid cluttering. So finally we have:

\begin{equation}
\small
\begin{aligned}
& \log p(\mathcal{R}|C,K) \\ 
\geq& \mathbb{E}_{R \in \mathcal{R}} \mathbb{E} \left[\log p(R|Z^s,Z^e)\right]\\
+& \mathbb{E}_{R \in \mathcal{R}} \mathbb{E} \left[\log p_\pi(Z^e|Z^s) +\log p_\pi(Z^s)-\log q_\phi(Z^s,Z^e)\right]
\end{aligned}    
\end{equation}

The likelihood term could be expanded as:
\begin{equation}
    \begin{aligned}
    &\mathbb{E}_{R \in \mathcal{R}} \mathbb{E}\left[ \log p_\theta(R|Z^s,Z^e) \right]\\
    =&\mathbb{E}_{R \in \mathcal{R}} \mathbb{E}\left[\prod\limits_{t=1}^{|l_R|} \log p_\theta(r_t|Z^s,Z^e,r_{<t})\right]\\
    \end{aligned}
\end{equation}

Note that the expectation above is with respect to the posterior $q(Z^s,Z^e)$. With mean-field approximation, we could assume that:
\begin{equation}
    q_\phi(Z^s,Z^e)=q_\phi(Z^s)q_\phi(Z^e)
\end{equation}
So the second term could be rewritten as:
\begin{equation}
\small
\begin{aligned}
&\mathbb{E}_{q_\phi(Z^s,Z^e)}[\log p_\pi(Z^e|Z^s)+\log p_\pi(Z^s)-\log q_\phi(Z^s,Z^e)]\\
=&\mathbb{E}_{q_\phi(Z^s)}[\log p_\pi(Z^s)-\log q_\phi(Z^s)]\\
+&\mathbb{E}_{ q_\phi(Z^s)}[\mathbb{E}_{q_\phi(Z^e)}\log p_\pi(Z^e|Z^s)-\log q_\phi(Z^e)]\\
=&{\rm{KL}}(q_\phi(Z^s)||p_\pi(Z^s))-\mathbb{E}_{ q_\phi(Z^s)}{\rm{KL}}(q_\phi(Z^e)||p_\pi(Z^e|Z^s))
\end{aligned}
\end{equation}

\section{Dataset Collection and Quality Control}
\label{sec:dataset}
Conversations in Reddit follow the pattern of ``initialize new topic-comment-reply-new comment-new reply'', and is suitable to construct a multi-reference knowledge-grounded conversation dataset in nature. A message tree is parsed for every post and every utterance is a node whose parent node is the comment it replies to and the root node is the initial utterance of the post host. A node and its all siblings are then viewed as multi-reference for a dialogue whose utterance history is the path from the root node to its parent node. The knowledge is crawled from a website whose URL is provided by the initial post.
Elaborated cleaning and filtering are conducted to ensure the quality of the dataset: (1) The length of response is no less than 6 tokens; (2) Only the knowledge sentence tagged as a paragraph in the website source code is kept; (3) The knowledge sentence in (2) should contain more than 15 tokens; (4) $\max\limits_{1 \leq i \leq n, 1 \leq j\leq m}Sim(R_i,K_j)\geq 0.1 $ and $n \geq 2$, $m \geq 3$ where $n$, $m$ are the number of responses and the number of knowledge sentences in a case, respectively. The similarity function is implemented as the unigram F1, which is coincident with the tagging of the pseudo span label. The date of collected Reddit conversations ranges from 2011 to 2017 following~\cite{qin-etal-2019-conversing}. The split of the dataset is based on date: January to June in 2012 for validation, July to December in 2012 for test, and the rest for training. This test set is referred to as General Test in the main document. 

To harvest the Focused Test, the filtering process is more sophisticated. Except from the aforementioned rules, we require that: (5) $ \mathop{\arg\max}\limits_{j}Sim(R_i,K_j)=j_0 \quad \forall i \in \{1,2,\cdots,n\}$. $n$ is the number of responses in a case and $j_0$ is the index of the collaborative attended knowledge. It means that all responses in a case are most similar to a single knowledge sentence, a much more challenging situation for a knowledge-grounded conversation model. However, using the lexical match to determine the groundings of the response is inaccurate. As a possible remedy, we hire  Amazon Mechanical Turk\footnote{\scriptsize\url{https://mturk.com}} annotators from native English-speaking countries with approval rates higher than 95\%. Each case meeting the above 5 rules is distributed to three workers to examine whether the multiple responses in the dialogue are referred to the same knowledge sentences or not and the majority of the labels are taken as the final decision. After the strict filtering and cleaning procedure, we finally get $833$ dialogues in the Focused Test.

\section{Implementation Details}
\label{sec:impl}

During the development of this paper, we adjust the learning rate from $1e-6$ to $1e-4$ and try batch sizes ranging from $16$ to $128$ and finally set the batch size to be $32$ since it produces the best result in the validation set. A cosine learning schedule is applied to adjust the learning rate during training. We set the minimum learning rate to be $0$ in the cosine learning schedule. The gradient clip is set to $2.0$ to avoid the explosion of the gradient. All modules are optimized with Adam with the hyper-parameters $\beta_1$=0.9, $\beta_2$=0.999.
When decoding, beam search is applied with a beam width of $2$. The length of the generated text is restrained in a range from $10$ to $30$. We set the repetition penalty to be $2.0$. 
The discriminator network and the grounding reward network use BERT$_{base}$ as backbone. The similarity score function $s(\cdot,\cdot)$ is implemented as unigram F1~\cite{dinan2018wizard} with the code shared at \url{https://github.com/facebookresearch/ParlAI/blob/master/parlai/core/metrics.py}. The hyper-parameter in training grounding reward network is $\mu=1$. We sweep the $\lambda$ and $\alpha$ from $1$ to $10$ and $0.5$ to $2.5$ respectively.
All experiments are performed on GTX 1080.

\section{Case Study}
\label{sec:case}
\begin{table*}[]
\resizebox{1.0\linewidth}{!}{
\begin{tabular}{c|l}
\thickhline
\multicolumn{2}{c}{Knowledge} \\ \hline
\multicolumn{2}{l}{...} \\
\multicolumn{2}{p{800pt}}{ Arnold, William. Fewer movies are being prescreened for critics and that's a good thing. Seattle Post Intelligencer. retrieved may 11, 2009.} \\
\multicolumn{2}{p{800pt}}{No movie shall triumph over \textit{Snakes on a Plane}. Unless I happen to feel like making a movie called more motherfucking snakes on more motherfucking planes.} \\
\multicolumn{2}{p{800pt}}{Snakes as bad as it wants to be, and that's good. Boston Globe. retrieved may 11, 2009.} \\
\multicolumn{2}{p{800pt}}{Samuel Jackson as agent Neville Flynn , an FBI agent assigned to protect Sean Jones on his flight to Los Angeles.} \\
\multicolumn{2}{l}{...} \\ \hline
\multicolumn{2}{c}{Context} \\ \hline
\multicolumn{2}{p{800pt}}{\textbf{U1}: That the working title for \textit{Snakes on a Plane} was changed to \textit{Pacific Air Flight 121} during filming, but Samuel Jackson insisted that they change it back. He later said, that's the only reason I took the job I read the title.} \\
\multicolumn{2}{p{800pt}}{\textbf{U2}: A much more interesting \textit{Snakes on a Plane} trivia fact is that it was in Sammy j's contract a snake couldn't come within 20 feet of him on set.}\\
\hline
\multicolumn{2}{c}{Response} \\ \hline
\multicolumn{1}{c|}{\multirow{3}{*}{SKT}}& \textbf{R1}: I'm not sure the movie was on a plane\\
 & \textbf{R2}: I'm not sure the movie was just a perfect example.\\
 & \textbf{R3}: I'm not sure the movie was just the one of the time.\\
 \hline
\multicolumn{1}{c|}{\multirow{3}{*}{VHRED$_{lgm}$}} & \textbf{R1}: I think the same thing that the movie was good.  \\
& \textbf{R2}: It's not a good movie. \\
& \textbf{R3}: I'm not sure if the movie is the same thing.\\
\hline
\multicolumn{1}{c|}{\multirow{3}{*}{Ours}} & \textbf{R1}:\textit{Snakes on a Plane} is a great movie, but I think it's more of a joke.\\
& \textbf{R2}: I think it's more likely that the snakes on the plane were actually snakes.\\
& \textbf{R3}: I think it was the best script for a movie.\\
\hline
\multicolumn{1}{c|}{\multirow{2}{*}{Human}} & \textbf{R1}: Was he scared of them or something? Seems like an odd clause for a movie like that.\\
& \textbf{R2}: And Samuel's lawyer rejoiced, having been given the opportunity to bill many hours for pleasing his whimsical client.\\
\thickhline

\end{tabular}
}
\caption{A case from General Test.}
\label{tab:case}

\end{table*}

Table \ref{tab:case} shows an example from the General Test. From the case we could have an intuitive cognition about the superiority of our model over existing VAE-based methods. The proposed model is not only expert in broadly concentrating on different knowledge sentences but also good at discovering ample semantics within a single knowledge sentence. Thus it is competent in generating diverse and knowledgeable responses. In contrast, the responses given by SKT and $\mathrm{VHRED}_{lgm}$ are either bland or tedious in semantics.

\end{document}